\documentclass[conference]{IEEEtran}
\IEEEoverridecommandlockouts

\usepackage{cite}
\usepackage{amsmath,amssymb,amsfonts}
\usepackage{algorithmic}
\usepackage{graphicx}
\usepackage{textcomp}
\usepackage{xcolor}
\usepackage{tikz}
\usetikzlibrary{arrows.meta,positioning}
\usepackage{booktabs}
\usepackage{placeins}
\usepackage[caption=false,font=footnotesize]{subfig}
\usepackage{eso-pic}

\usepackage{doi}

\usepackage[nameinlink,capitalise,noabbrev]{cleveref}
\crefname{figure}{Fig.}{Figs.}
\Crefname{figure}{Fig.}{Figs.}
\crefname{table}{Table}{Tables}
\Crefname{table}{Table}{Tables}
\crefname{equation}{Eq.}{Eqs.}
\Crefname{equation}{Eq.}{Eqs.}

\def\BibTeX{{\rm B\kern-.05em{\sc i\kern-.025em b}\kern-.08em
    T\kern-.1667em\lower.7ex\hbox{E}\kern-.125emX}}

\usepackage{fancyhdr}
\fancypagestyle{firstpage}{
    \fancyhf{} 
    \fancyhead[L]{Accepted at 2025 28th International Conference on Computer and Information Technology (ICCIT)\\
    19--21 December 2025, Cox’s Bazar, Bangladesh}

}

\begin{document}

\AddToShipoutPictureFG*{%
  \AtPageUpperLeft{%
    \put(52,-23){%
      \parbox[t]{\dimexpr\paperwidth-104pt\relax}{%
        \fontsize{8}{9}\selectfont
        2026 29th International Conference on Computer
        and Information Technology (ICCIT)\\
        18--20 December 2026, Cox's Bazar, Bangladesh
      }%
    }%
  }%
}

\title{Beyond Cross-Modal Reconstruction: Participant-Level Evaluation of sEMG–Kinematic Coupling in Post-Stroke Gait}

\author{

\IEEEauthorblockN{Rafsan Jany}
\IEEEauthorblockA{
Digital Health Research Division, KIOM\\
University of Science and Technology\\
Daejeon, Republic of Korea\\
rafsanjany@ust.ac.kr
}

\and

\IEEEauthorblockN{Md Rafid Islam}
\IEEEauthorblockA{
Electrical and Computer Engineering\\
North South University\\
Dhaka, Bangladesh\\
md.islam.241@northsouth.edu
}

\and

\IEEEauthorblockN{Mohammad Fahim Islam}
\IEEEauthorblockA{
Computer Science and Engineering\\
American International University-Bangladesh\\
Dhaka, Bangladesh\\
23-51971-2@student.aiub.edu
}

\and

\IEEEauthorblockN{Hafiz Abdur Rahman\textsuperscript{*}}
\IEEEauthorblockA{
Electrical and Computer Engineering\\
North South University\\
Dhaka, Bangladesh\\
hafiz.rahman@northsouth.edu\\
\textsuperscript{*}Corresponding Author
}
}
\maketitle

\begin{abstract}
Accurate cross-modal reconstruction is often treated as evidence of physiological coupling, although it may simply reflect temporal patterns shared across participants. We tested whether frozen bidirectional models captured subject-specific neuromuscular coupling between electromyographic activity and gait kinematics in 82 held-out able-bodied and 43 post-stroke participants. Complete source-modality removal and cross-participant source substitution, along with a cycle-position-age-sex template baseline, were applied. Learned bilateral asymmetry was also compared with transparent waveform and derivative asymmetry for predicting Performance-Oriented Mobility Assessment (POMA) scores in the case of repeated nested cross-validation. Complete source removal increased reconstruction RMSE by 6.90\% for electromyography-to-kinematics and 11.38\% in the reverse direction. In contrast, cross-participant substitution increased RMSE by only 1.68\% and 1.67\%, respectively, while the template reduced RMSE by 2.89\% and 0.55\%. For POMA prediction, walking speed, age, and sex achieved an MAE of 3.693, compared with 4.229 and 4.427 after adding absolute and signed learned asymmetry, respectively. In all of the 20 repeated evaluations, all asymmetry representations performed worse than the covariate baseline. No localized learned feature remained significant after covariate adjustment and false-discovery-rate correction. The model mainly relied on shared gait-phase patterns, with no clear subject-specific coupling or added clinical value. Participant-level controls are therefore needed before reconstruction accuracy is treated as evidence of individualized physiology.
\end{abstract}

\begin{IEEEkeywords}
Gait analysis, cross-modal reconstruction, neuromuscular coupling, post-stroke gait, electromyography, kinematics, temporal convolutional networks, bilateral asymmetry, POMA, biomarker validation, falsification framework.
\end{IEEEkeywords}

\section{Introduction}

Walking after a stroke is not simply a healthy gait performed more slowly. Muscle coordination may be reorganized, bilateral timing becomes uneven, and muscle activation and joint motion can change in different ways across individuals~\cite{Clark2010,Patterson2010}. Surface electromyography (sEMG) reflects the timing and intensity of superficial muscle activation; kinematics describes the resulting movement. Examining these signals together may therefore reveal aspects of gait organization which could be missed by either modality alone. This opportunity is supported by public resources such as the dataset of Van Criekinge \textit{et al.}, which provides synchronized gait recordings from 138 able-bodied adults and 50 stroke survivors~\cite{VanCriekinge2023}. Recent studies using this resource have reported stroke classification and waveform-based deviation or asymmetry measures~\cite{Romano2024,Calabro2026}.

Cross-modal neural networks offer an appealing way to connect muscle activity with movement. Deep models have achieved accurate joint-kinematic prediction from sEMG and have also incorporated biomechanical constraints into muscle-to-motion estimation~\cite{Song2023,Zhang2023}. These results make reconstruction error a convenient measure of model performance, but its physiological interpretation is less straightforward. A low reconstruction error shows that one signal can help predict another; it does not, by itself, show that the model has learned an individualized neuromuscular relationship.

This distinction is especially important for time-normalized gait cycles. Cycle position, population-average waveform structure, and demographic information may already explain a substantial part of a target waveform. A model can exploit such regularities while appearing successful on a conventional test set, a behavior closely related to shortcut learning~\cite{Geirhos2020}. Participant-level separation is essential when repeated observations are available because record-wise validation can substantially overestimate clinical performance~\cite{Saeb2017}. However, even a participant-disjoint evaluation does not establish individualized coupling unless the model is shown to require the correct participant's source signal.

We evaluate 82 untouched able-bodied participants and 43 stroke participants using three complementary controls with frozen, healthy-trained BioCoupleNet models. These controls are complete source-modality removal, cross-participant source substitution, and a cycle-position--age--sex template without opposite-modality input. We then compare learned bilateral asymmetry with transparent waveform and derivative asymmetry and test whether any representation improves prediction of the Performance-Oriented Mobility Assessment (POMA) beyond walking speed, age, and sex~\cite{Tinetti1986}. The results show that the model uses general phase-aligned information, but neither reconstruction nor learned asymmetry establishes subject-specific coupling or additional clinical value. This study provides a participant-level falsification framework and a practical message, and that is reconstruction accuracy should not be treated as physiological validity unless the correct participant's source signal is demonstrably necessary.

\section{Dataset and Analysis Cohort}

We used the post-processed MATLAB release of the public full-body gait dataset introduced by Van Criekinge \textit{et al.}~\cite{VanCriekinge2023,VanCriekingeFigshare2023}. The original dataset contains 138 able-bodied adults and 50 stroke survivors walking barefoot at their preferred speed along a 12-m walkway. It provides synchronized full-body kinematics and bilateral surface electromyography (sEMG). We selected seven ipsilateral sEMG channels---gastrocnemius (GAS), rectus femoris (RF), vastus lateralis (VL), biceps femoris (BF), semitendinosus (ST), tibialis anterior (TA), and erector spinae (ERS)---together with sagittal hip, knee, ankle, and pelvis angles. The released sEMG signals were already rectified, filtered, smoothed, and normalized to each participant's maximum across the available strides. Therefore, the study focused on the timing and pattern of muscle activity, not the amount of muscle activation.

A stride was retained only when all seven sEMG and four kinematic signals were available. Waveforms containing no more than 5\% missing samples were linearly interpolated. The remaining incomplete strides were excluded. Each accepted waveform was resampled to 256 points over the normalized 0--100\% gait cycle. After quality control, the analysis included 82 able-bodied participants with 718 strides and 43 stroke participants with 644 strides. All stroke participants had complete paretic and nonparetic recordings, as it was a requirement for bilateral asymmetry analysis. For healthy representation learning, every signal-complete side was retained, because each side constituted a valid normative observation. Age, sex, and side-specific walking speed were available for both groups. POMA was used as the primary clinical outcome for the stroke cohort. All model development and evaluation splits were formed at the participant level. Additionally, it is necessary to note that healthy outer participants were not used for scaling, fitting, early stopping, or checkpoint selection. The original investigators reported informed consent and institutional ethical approval~\cite{VanCriekinge2023}.

\section{Proposed Framework}
\label{sec:framework}

\subsection{Framework Overview}

Earlier work has shown that muscle-level EMG features can support interpretable analysis of post-stroke gait~\cite{HussainJany2024}. Our question is narrower, which is, can a cross-modal model move beyond group-level discrimination and capture a participant-specific relationship between muscle activity and movement? To examine this, the framework has three stages, and these are healthy-only cross-modal learning, anatomically localized disruption scoring, and participant-level physiological validity testing.

\begin{figure}[htbp]
\centering
\includegraphics[width=\columnwidth]{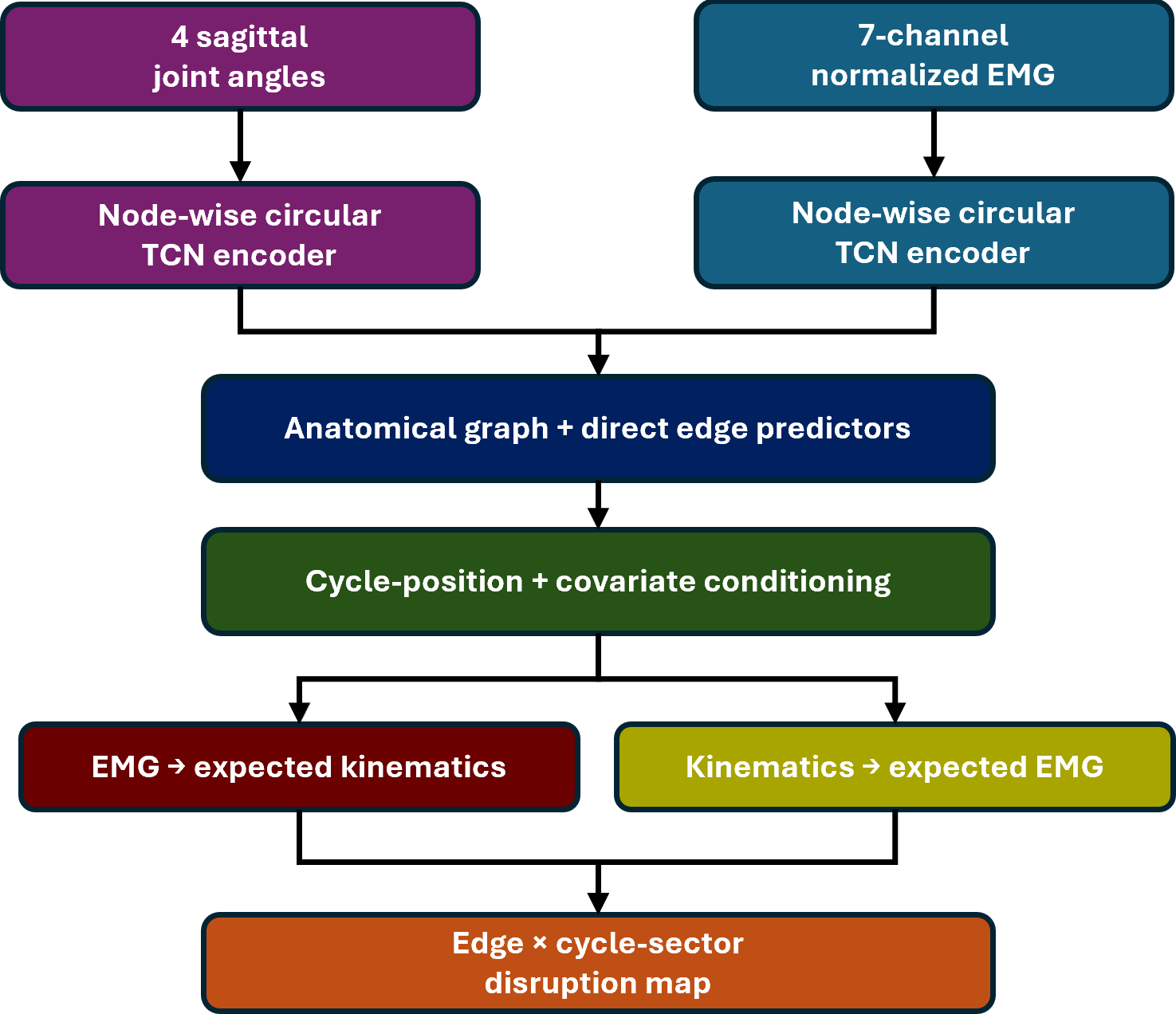}
\caption{Architecture of the frozen BioCoupleNet model. Separate circular TCN encoders represent sEMG and kinematic waveforms. Direct predictors operate only on prespecified muscle--joint connections and reconstruct the opposite modality in both directions. During stroke inference, bidirectional edge residuals are summarized over normalized gait-cycle sectors and converted into paretic--nonparetic asymmetry. Stroke outcomes are not used to train the representation.}
\label{fig:biocouplenet_framework}
\end{figure}

BioCoupleNet learns the expected relationship between seven sEMG waveforms and four sagittal kinematic waveforms from healthy gait only. The frozen model is then applied to paretic and nonparetic gait without using stroke labels or clinical outcomes. Its residuals are converted into bilateral asymmetry features and, after that, tested against transparent signal-based baselines and participant-level falsification controls. The architecture of the frozen BioCoupleNet model is presented in Fig.~\ref{fig:biocouplenet_framework}.

\subsection{Anatomy-Constrained Cross-Modal Model}

For stride \(i\), the model receives the sEMG matrix \(\mathbf{E}_i\in\mathbb{R}^{7\times T}\), the kinematic matrix \(\mathbf{K}_i\in\mathbb{R}^{4\times T}\), and robustly scaled age and sex covariates. Each signal node is encoded by a shared temporal convolutional network with five residual blocks and dilations \(1,2,4,8,\) and \(16\). Following the general TCN design \cite{Bai2018}, the residual blocks preserve temporal resolution and provide a broad receptive field. We use circular rather than zero padding because each input is a complete normalized gait cycle. The beginning and end of the waveform are treated as adjacent points.

A learned node embedding retains the identity of each muscle or joint. Continuous cycle position is represented by

\begin{equation}
\boldsymbol{\phi}(t)=
\left[
\sin(2\pi t),\,
\cos(2\pi t),\,
\sin(4\pi t),\,
\cos(4\pi t)
\right],
\end{equation}

where \(t\in[0,1)\). This encoding, together with age and sex, conditions the
edge predictors without requiring imputed internal gait events.

The graph contains ten prespecified connections, and these are GAS and TA to the ankle, RF to the hip and knee, VL to the knee, BF and ST to the hip and knee, and ERS to the pelvis. For every permitted edge \((m,j)\), the model predicts the joint waveform from muscle \(m\), denoted \(\widehat{\mathbf{K}}^{\,m\rightarrow j}\), and predicts the muscle waveform from joint \(j\), denoted \(\widehat{\mathbf{E}}^{\,j\rightarrow m}\). Predictions from edges sharing a target are combined using learned softmax weights. The complete model contains 45,372 trainable parameters.

\subsection{Healthy-Only Learning Objective}

Training uses only healthy participants and is performed independently within each participant-disjoint fold. One input channel is masked with probability \(0.20\) and a contiguous segment covering \(12\%\) of the gait cycle is masked with probability \(0.35\). The goal is to discourage direct copying. The same masking policy is applied independently to both modalities.

For each reconstruction direction \(d\), the loss combines a sector-balanced Smooth-\(L_1\) term, waveform correlation, and first-derivative agreement:

\begin{equation}
\mathcal{L}_{d} =
\mathcal{L}^{d}_{\mathrm{SL1}}
+0.15\mathcal{L}^{d}_{\mathrm{corr}}
+0.10\mathcal{L}^{d}_{\Delta}.
\end{equation}

The Smooth-\(L_1\) loss is calculated separately within four equal normalized cycle sectors and then averaged, preventing one part of the gait cycle from dominating training. The complete objective is

\begin{equation}
\mathcal{L} =
\mathcal{L}_{E\rightarrow K}
+\mathcal{L}_{K\rightarrow E}
+0.15\mathcal{L}_{\mathrm{self}}
+0.35\mathcal{L}_{\mathrm{edge}},
\end{equation}

where \(\mathcal{L}_{\mathrm{self}}\) regularizes within-modality reconstruction and \(\mathcal{L}_{\mathrm{edge}}\) directly supervises every allowed muscle--joint predictor. POMA and all other stroke outcomes remain outside model training, early stopping, and checkpoint selection.

\subsection{Normative Disruption and Bilateral Asymmetry}

For edge \((m,j)\) and time point \(t\), the bidirectional residual is
\begin{align}
r_{i,mj}(t)
= \frac{1}{2}\Big[&
0.7\left|
\widehat{K}^{,m\rightarrow j}_{i}(t)-K_{i,j}(t)
\right| \nonumber\\
&+0.3\left|
\Delta\widehat{K}^{,m\rightarrow j}_{i}(t)
-\Delta K_{i,j}(t)
\right| \nonumber\\
&+0.7\left|
\widehat{E}^{,j\rightarrow m}_{i}(t)-E_{i,m}(t)
\right| \nonumber\\
&+0.3\left|
\Delta\widehat{E}^{,j\rightarrow m}_{i}(t)
-\Delta E_{i,m}(t)
\right|
\Big]
\end{align}

Residuals are averaged within each of the four cycle sectors. For edge-sector pair \(e\), the resulting score \(R_{i,e}\) is standardized using the post-training healthy reference participants from the corresponding outer fold:
\begin{equation}
z_{i,e}=
\frac{R_{i,e}-\operatorname{median}(R_{h,e})}
{\max\!\left(1.4826\,\operatorname{MAD}(R_{h,e}),10^{-4}\right)}
\end{equation}

Over the strides, the standardized values are averaged and summarized into six prespecified modules. These modules are GAS--ankle mid-cycle; TA--ankle late-cycle; early-cycle knee extensors; late-cycle hamstring-knee coupling; hip-muscle coupling; and full-cycle ERS-pelvis coupling. For participant \(i\) and module \(m\), the signed and absolute bilateral representations are
\begin{equation}
A^{\mathrm{signed}}_{i,m}
=M^{P}_{i,m}-M^{N}_{i,m},
\qquad
A^{\mathrm{absolute}}_{i,m}
=\left|M^{P}_{i,m}-M^{N}_{i,m}\right|
\end{equation}
where \(P\) and \(N\) denote the paretic and nonparetic sides.

Instead of being considered as physiological biomarkers by construction, these scores are regarded as candidate representations. While primarily depending on cycle position or population waveform structure, a model may produce stable residuals~\cite{Geirhos2020}. The following experimental protocol tests whether the correct participant's source modality is genuinely necessary through complete-source removal, cross-participant substitution, and a cycle-position--age--sex template. These controls are applied only after the model is frozen and do not guide architecture selection.

\begin{figure*}[t]
\centering
\includegraphics[width=0.8\textwidth]{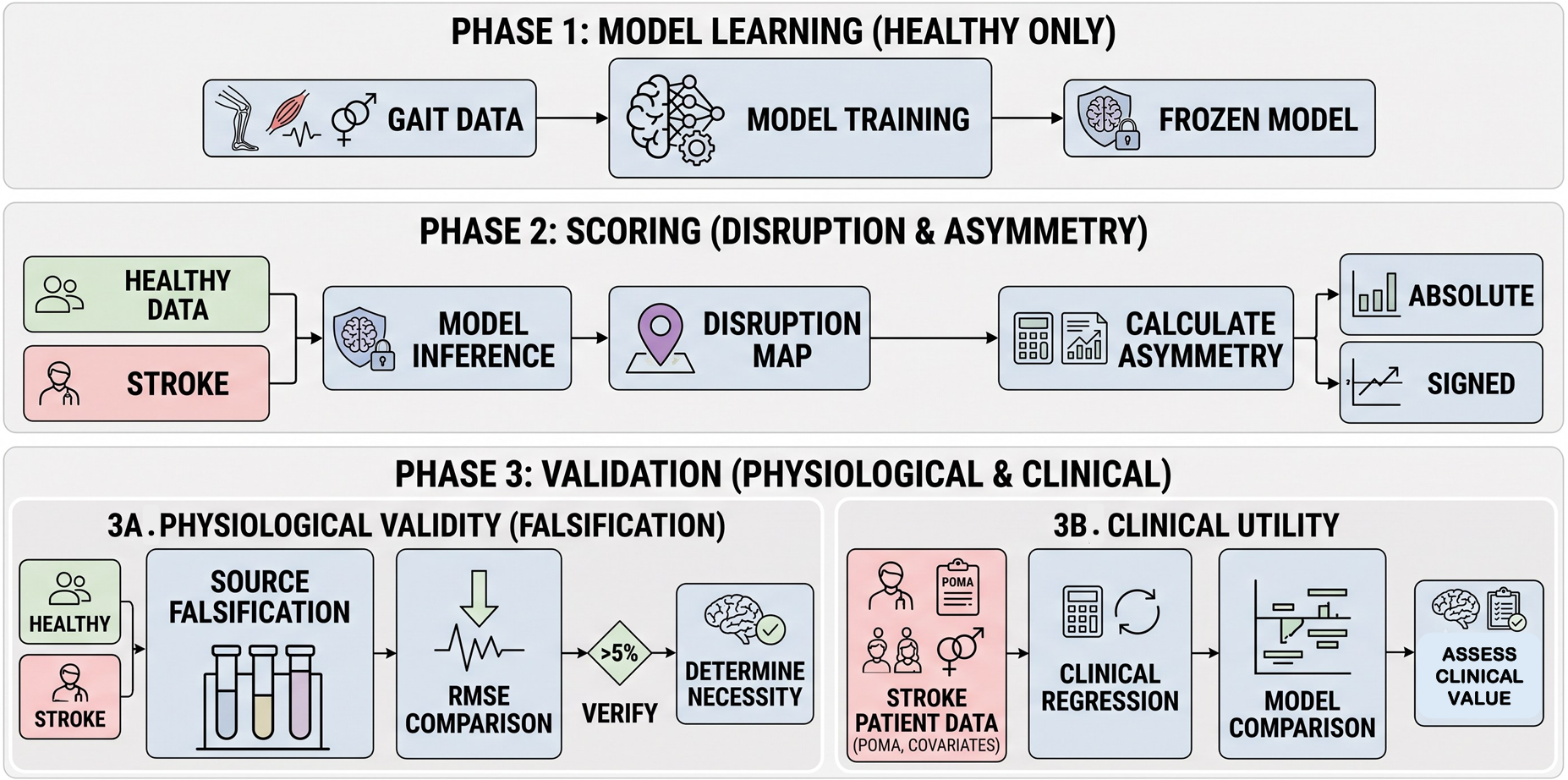}
\caption{Participant-level evaluation of cross-model gait coupling.}
\label{fig:workflow}
\end{figure*}

The experimental process is illustrated in Fig.~\ref{fig:workflow}.

\section{Results and Discussion}
\label{sec:results}

\subsection{Evaluation Protocol}

Each reconstruction experiment was evaluated using three seeds and five participant-disjoint outer folds. The final falsification analysis reused all 15 frozen BioCoupleNet checkpoints. Its primary cohort comprised 82 outer-test healthy participants. The 43 stroke participants were examined as a transfer cohort. A control was considered physiologically meaningful when two conditions were met. These conditions are an increased participant-level RMSE by at least the prespecified 5\% threshold and remaining significant after false-discovery-rate (FDR) correction. Confidence intervals were estimated using 5,000 participant-level bootstrap samples, and paired inference used 10,000 sign-flip repetitions. The architecture, controls, and thresholds were frozen before the final analysis.

\subsection{Reconstruction Performance Was Encouraging but Incomplete}

At first sight, the reconstruction results were encouraging. BioCoupleNet achieved mean held-out kinematic and sEMG RMSE values of 0.517 and 0.667, with corresponding waveform correlations of 0.771 and 0.721. Replacing the anatomical graph with a fully connected graph increased these RMSE values to 0.538 and 0.695. Removing cycle position produced the largest deterioration, increasing kinematic RMSE to 0.576 and sEMG RMSE to 0.706. This corresponds to relative increases of approximately 11.4\% and 5.9\%, respectively.

However, removing the direct edge loss produced almost unchanged results. Kinematic and sEMG RMSE remained 0.518 and 0.667, respectively. Similarly, masking any single sEMG channel changed kinematic RMSE by no more than 1.14\%. These findings suggest that anatomical restriction was modestly helpful, but cycle position provided the strongest predictive cue. They also raised a more difficult question, and that is, was the model reconstructing participant-specific muscle--movement relationships, or was it mainly using the shared structure of a normalized gait cycle?

The learned representation was numerically stable. Pairwise correlations between seed-specific disruption maps ranged from 0.974 to 0.983, and paretic and nonparetic split-half correlations were 0.976 and 0.952. Stability alone, however, does not establish physiological validity. A consistently learned shortcut remains a shortcut~\cite{Geirhos2020}.

\subsection{Participant-Level Falsification}

\begin{figure*}[t]
    \centering
    \includegraphics[width=\textwidth]
    {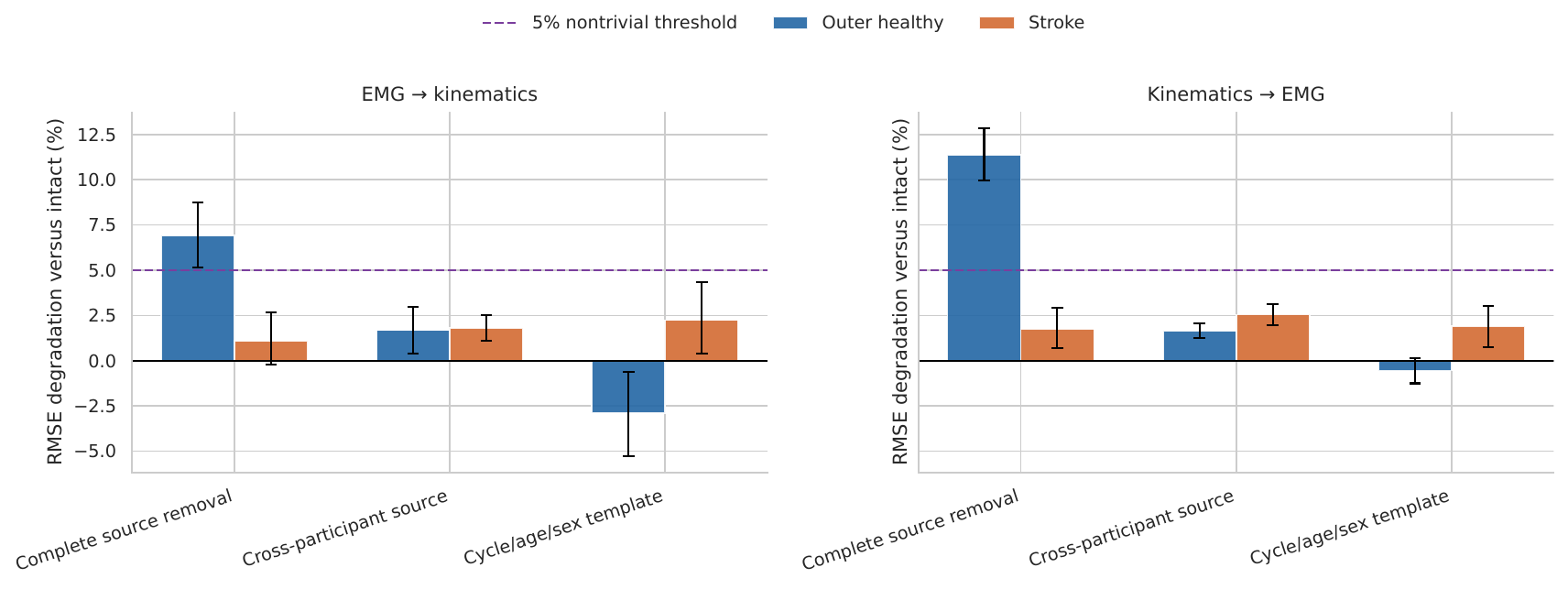}
    \caption{How reconstruction error changed when each falsification control was applied. The bars report the mean RMSE difference from the intact model, and the whiskers give the bootstrap 95\% confidence intervals. We chose a 5\% change in advance as the cutoff for a meaningful effect. Only removing the source signal passed this cutoff in the held-out healthy group. Using another participant's signal or the cycle-position--age--sex template did not. Negative values favor the control.}
    \label{fig:cross_modal_falsification}
\end{figure*}

Fig.~\ref{fig:cross_modal_falsification} presents the decisive source-dependence tests. On the untouched healthy cohort, removing the complete source modality increased RMSE by 6.90\% for sEMG-to-kinematics and 11.38\% for kinematics-to-sEMG. Both effects exceeded the 5\% threshold, with bootstrap 95\% CIs of [5.15, 8.72]\% and [9.97, 12.82]\%, respectively, and both remained significant after FDR correction ($p_{\mathrm{FDR}}=0.0002$). Thus, the model did use information from the source modality.

Individualized coupling was not supported by the more informative controls. RMSE increased by just 1.68\% for sEMG-to-kinematics and 1.67\% in the opposite direction when each participant's source signal was substituted with another participant's signal. Although their confidence intervals stayed below the frozen 5\% threshold, these changes were statistically noticeable. The difference is important since the effect-size gate asks whether the correct participant's signal caused a meaningful difference, whereas a small $p$-value indicates that the change was consistent.

Even more insightful was the cycle-position--age--sex template. It generated a minor 0.55\% reduction for sEMG reconstruction and a 2.89\% drop in kinematic RMSE when compared to intact BioCoupleNet. Therefore, without getting the opposite modality, the template matched or performed better than the cross-modal network. The highest decline in the stroke cohort was 2.55\%, and none of the three controls crossed the 5\% threshold in either direction.

Taken together, these results support a narrow conclusion. BioCoupleNet used general phase-aligned source information because removing the source entirely was harmful. Nevertheless, it did not require enough participant-specific information to establish individualized neuromuscular coupling. Cross-participant substitution is especially important here because conventional participant-disjoint testing prevents identity leakage but does not show that the correct participant's source signal is necessary~\cite{Saeb2017}.

\subsection{Group Differences Were Not Clinically Specific}

The initial disruption scores showed large differences between healthy and stroke gait. Median normative disruption was 0.128 in held-out healthy participants, compared with 2.100 on the paretic side and 1.988 on the nonparetic side. The corresponding Cliff's delta values were 0.892 and 0.928, with $p_{\mathrm{FDR}}<10^{-15}$ in both comparisons. These are strong group-level differences.

They should not, however, be interpreted as localized paretic-side coupling abnormalities. Paretic and nonparetic total disruption did not differ significantly ($p=0.105$), and both sides showed similar increases relative to healthy gait. Moreover, the healthy 2.5--97.5\% walking-speed range was 0.909--1.567~m/s, and only 3 of the 43 stroke participants fell within this range. The observed separation may therefore reflect a mixture of stroke status, walking speed, general gait severity, and cross-domain waveform differences. The current cohort cannot separate these explanations reliably.

\subsection{No Incremental Clinical Value}

\begin{figure}[htbp]
\centering
\includegraphics[width=\columnwidth]{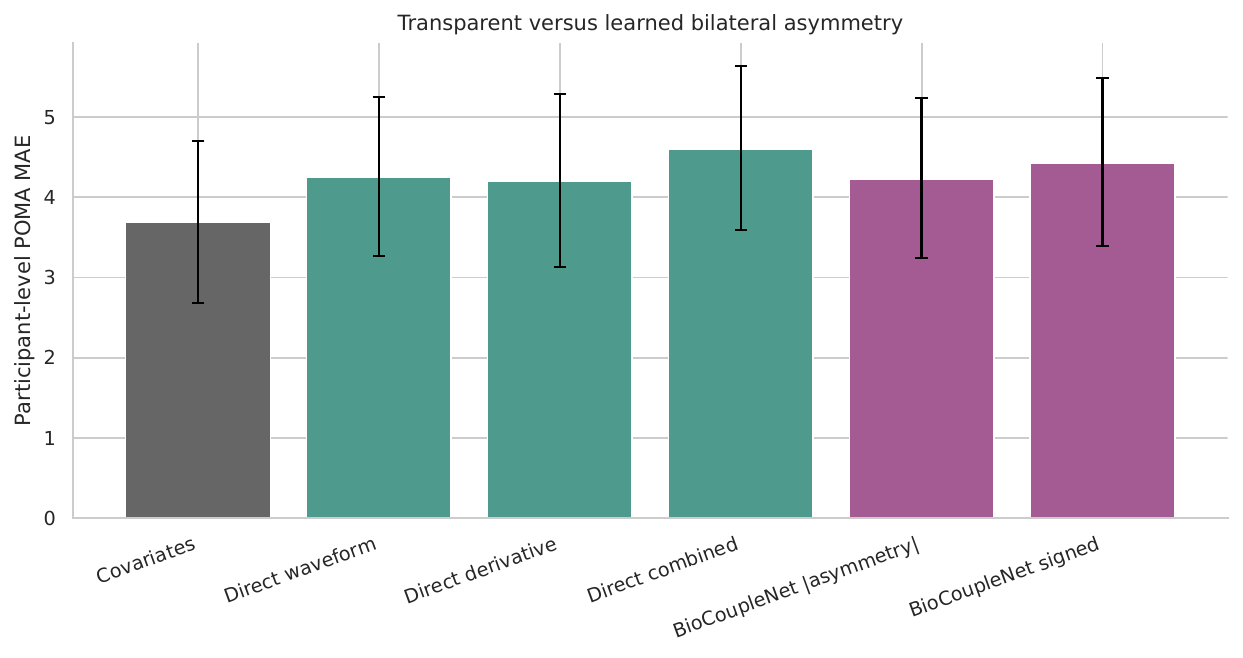}
\caption{Participant-level POMA benchmark for transparent and learned
    bilateral asymmetry representations. Bars show nested-CV MAE and error
    bars show participant-bootstrap 95\% confidence intervals. The
    walking-speed--age--sex baseline achieved the lowest error. Neither the
    transparent signal measures nor the BioCoupleNet representations added
    predictive value.}
\label{fig:clinical_benchmark}
\end{figure}

Table~\ref{tab:poma_benchmark} and Fig.~\ref{fig:clinical_benchmark} compare the learned representation with transparent asymmetry measures under identical participant-level nested validation. Walking speed, age, and sex alone achieved the lowest POMA MAE of 3.693 and a Spearman correlation of 0.583.

\begin{table*}[t]
\caption{Repeated participant-level nested-CV prediction of POMA.
All asymmetry methods include walking speed, age, and sex. Lower MAE is
better. Values in brackets are participant-bootstrap 95\% confidence
intervals.}
\label{tab:poma_benchmark}
\centering
\footnotesize
\begin{tabular}{@{}lccc@{}}
\toprule
\textbf{Representation} &
\textbf{Features} &
\textbf{MAE [95\% CI]} &
\textbf{Spearman $\rho$ [95\% CI]} \\
\midrule
Walking speed + age + sex
    & 3  & \textbf{3.693 [2.685, 4.699]}
    & \textbf{0.583 [0.312, 0.812]} \\
Direct derivative asymmetry + covariates
    & 9  & 4.205 [3.132, 5.284]
    & 0.518 [0.212, 0.768] \\
BioCoupleNet absolute asymmetry + covariates
    & 9  & 4.229 [3.239, 5.237]
    & 0.545 [0.272, 0.756] \\
Direct waveform asymmetry + covariates
    & 9  & 4.249 [3.264, 5.250]
    & 0.557 [0.292, 0.791] \\
BioCoupleNet signed asymmetry + covariates
    & 9  & 4.427 [3.393, 5.480]
    & 0.440 [0.120, 0.719] \\
Direct waveform + derivative asymmetry + covariates
    & 15 & 4.598 [3.596, 5.634]
    & 0.489 [0.194, 0.737] \\
\bottomrule
\end{tabular}
\end{table*}

Adding BioCoupleNet absolute asymmetry increased mean absolute error by 0.536 POMA points relative to the covariate baseline, with a paired 95\% CI of [0.013, 1.079]. Signed asymmetry increased error by 0.734 points, with a 95\% CI of [0.417, 1.043]. The absolute and signed representations worsened participant-level prediction for 26 and 32 of the 43 participants, respectively. More broadly, every asymmetry representation was worse than the covariate baseline across all 20 repeated evaluations.

The learned representation also failed to outperform the transparent alternatives. Direct derivative asymmetry achieved an MAE of 4.205, compared with 4.229 for BioCoupleNet absolute asymmetry. Although the learned representation performed better than the combined waveform--derivative representation, their paired error difference was not significant (95\% CI [$-$0.978, 0.237], $p_{\mathrm{FDR}}=0.753$). The network therefore did not provide evidence of information that could not be obtained from simpler signal calculations.

No BioCoupleNet module remained associated with POMA after adjusting for walking speed, age, and sex and applying FDR correction. The smallest adjusted FDR value among the learned modules was 0.366. Direct derivative GAS--ankle and hip-muscle features showed the strongest exploratory trends, but both had $p_{\mathrm{FDR}}\approx0.092$ and cannot be treated as confirmed associations.

\subsection{Scientific Meaning and Limitations}

The significance of this work is more in what the controls show than in the particular network. Large patient-control differences, reproducible maps, and accurate reconstruction may seem convincing, but they do not prove that a model learned the required participant-specific physiology. Successful cross-modal learning would have been implied by complete source removal alone. That interpretation was altered by the phase-only template and cross-participant substitution.

This study does not show that neuromuscular coupling is absent after stroke. It shows that BioCoupleNet reconstruction and residual asymmetry did not establish coupling in this cohort. It also does not support diagnosis, recovery prediction, treatment-response assessment, rehabilitation prescription, or a validated clinical biomarker.

Only 43 stroke patients were included in the clinical study, and there was minimal walking-speed overlap with the pool of healthy individuals. Analysis of the absolute activation magnitude was not possible since the emitted sEMG waveforms were normalized. It was not possible to estimate the trial-level ICC either, since genuine repeated-trial identifiers were not available. Lastly, there was no external, speed-overlapping validation cohort; all evidence came from a single public dataset. The main methodological finding, that physiological interpretation requires proof that the right participant's source signal is required, not just reconstruction accuracy, is not compromised by these restrictions, although they do impede a clinical claim.

\section{Conclusion}
\label{sec:conclusion}

This study asked whether accurate cross-modal reconstruction is sufficient evidence that a neural network has learned individualized neuromuscular coupling. BioCoupleNet produced stable bidirectional reconstructions, and complete source removal increased RMSE by 6.90\% and 11.38\%. This is a confirmation that the model used information from the opposite modality. However, replacing the correct participant's source signal with another participant's signal increased RMSE by only about 1.7\%, while a cycle-position--age--sex template matched or outperformed the intact model. Learned asymmetry also failed to improve POMA prediction beyond walking speed, age, and sex.

The outcome of the experiment is helpful yet cautious. It indicates that reconstruction accuracy and stable residual patterns were insufficient to establish a neuromuscular connection in this model and population, not that it is lacking after stroke. Therefore, the primary contribution is a practical validation principle. The principle is that physiological claims should be supported by evidence that the source signal from the right individual is required and that the resulting representation provides value beyond simple clinical factors. This criterion should be tested in larger, externally collected, speed-matched cohorts in future research.

\bibliographystyle{IEEEtran}
\bibliography{references}

\end{document}